%% file: main.tex
\documentclass{llncs}
\usepackage[american]{babel}
\usepackage[T1]{fontenc}
\usepackage{times}
\usepackage{graphicx}

\usepackage{amsmath,amssymb,amsfonts}   
\usepackage[ruled, linesnumbered,lined,boxed]{algorithm2e}
\usepackage{xcolor}
\usepackage{hyperref}
\usepackage[acronym]{glossaries}
\glsdisablehyper

\newcommand{\norm}[1]{\left\| #1 \right\|_2}

\newcommand{\loss}{\ensuremath{\mathcal{L}}}
\definecolor{ao}{rgb}{0.0, 0.5, 0.0}
\definecolor{babyblueeyes}{rgb}{0.63, 0.79, 0.95}
\definecolor{brickred}{rgb}{0.8, 0.25, 0.33}
\definecolor{bronze}{rgb}{0.8, 0.5, 0.2}
\definecolor{darkgreen}{rgb}{0.0, 0.5, 0.0}

\definecolor{rub_orange}{RGB}{195, 139, 4}
\definecolor{rub_green}{RGB}{30,110,40}
\definecolor{rub_red}{RGB}{200, 0, 0}

\definecolor{rub_blue}{RGB}{0,53,96}
\definecolor{rub_gray}{RGB}{1,1,1}
\definecolor{bg_white}{RGB}{255,255,255}

\usepackage[most]{tcolorbox}

\usepackage{xparse}

\NewDocumentEnvironment{testexample}{ O{} }
{
\colorlet{colexam}{black} 
\newtcolorbox[use counter=testexample]{testexamplebox}{%
    empty,
    size=minimal,
    title={\small Fanfiction excerpt~\thetcbcounter : #1},
    attach boxed title to top left,
       minipage boxed title,
    boxed title style={empty,size=minimal,toprule=0pt,top=2pt,left=1.3mm,overlay={}}, 
    coltitle=colexam,fonttitle=\bfseries,
    before=\par\medskip\noindent,parbox=false,boxsep=0pt,left=1.3mm,right=0mm,top=2pt,
       before upper=\csname @totalleftmargin\endcsname0pt, 
    overlay unbroken={\draw[colexam,line width=.75pt] ([xshift=1.6pt, yshift=-1.2pt]title.north west) -- ([xshift=1.6pt, yshift=-1.2pt]frame.south west); },
    }
\begin{testexamplebox}}
{\end{testexamplebox}\endlist}

\newacronym{LEV}{LEV}{\textit{linguistic embedding vector}}
\newacronym{PLDA}{PLDA}{\textit{probabilistic linear discriminant analysis}}
\newacronym{AV}{AV}{authorship verification}
\newacronym{LEVs}{LEVs}{\textit{linguistic embedding vectors}}

\usepackage{rotating}
\usepackage{multirow}

\usepackage{floatrow}
\newfloatcommand{capbtabbox}{table}[][\FBwidth]
\begin{document}

{\let\thefootnote\relax\footnotetext{\scriptsize Copyright \textcopyright\ 2020 for this paper by its authors. Use permitted under Creative Commons License Attribution 4.0 International (CC BY 4.0). CLEF 2020, 22-25 September 2020, Thessaloniki, Greece.}}

\setlength{\abovedisplayskip}{5.pt}
\setlength{\belowdisplayskip}{6.pt}
\setlength{\abovedisplayshortskip}{5.pt}
\setlength{\belowdisplayshortskip}{6.pt}

\title{Deep Bayes Factor Scoring for Authorship Verification}

\subtitle{Notebook for PAN 2020 at CLEF 2020}

\author{Benedikt Boenninghoff~\inst{1} \and Julian Rupp\inst{1}  \and Robert M. Nickel\inst{2} \and Dorothea Kolossa\inst{1}}
\institute{Ruhr University Bochum, Germany \\
            \email{\{benedikt.boenninghoff, julian.rupp, dorothea.kolossa\}@rub.de}
            \and
           Bucknell University, Lewisburg, PA, USA \\
        \email{rmn009@bucknell.edu}
}
\maketitle

\begin{abstract}
\vspace*{-0.4cm}

The PAN 2020 authorship verification (AV) challenge focuses on a cross-topic/closed-set AV task over a collection of fanfiction texts. Fanfiction is a fan-written extension of a storyline in which a so-called fandom topic describes the principal subject of the document. The data provided in the PAN 2020 AV task is quite challenging because authors of texts across multiple/different fandom topics are included. In this work, we present a hierarchical fusion of two well-known approaches into a single end-to-end learning procedure: A deep metric learning framework at the bottom aims to learn a pseudo-metric that maps a document of variable length onto a fixed-sized feature vector. At the top, we incorporate a probabilistic layer to perform Bayes factor scoring in the learned metric space. We also provide text preprocessing strategies to deal with the cross-topic issue.

\end{abstract}

\vspace*{-0.8cm}
\section{Introduction}
\vspace*{-0.15cm}
\input{sections/sec_1_intro}

\vspace*{-0.15cm}
\section{\textsc{AdHominem:} Siamese network for representation learning} 
\vspace*{-0.15cm}
\input{sections/sec_2_theory}

\vspace*{-0.15cm}
\section{Text preprocessing strategies} 
\vspace*{-0.15cm}
\input{sections/sec_3_implementation}

\vspace*{-0.25cm}
\section{Evaluation}
\vspace*{-0.15cm}
\input{sections/sec_4_results}

\vspace*{-0.25cm}
\section{Conclusion and future work}
\vspace*{-0.15cm}
\input{sections/sec_5_conclusion}

\vspace*{-0.25cm}
\section*{Acknowledgment}
\vspace*{-0.2cm}
This work was in significant parts performed on a HPC cluster at Bucknell University through the support of the National Science Foundation, Grant Number 1659397. Project funding was provided by the state of North Rhine-Westphalia within the Research Training Group "SecHuman - Security for Humans in Cyberspace." 

\vspace*{-0.25cm}
\bibliographystyle{splncs03}
\begin{raggedright}
\bibliography{refs}
\end{raggedright}

\end{document}

%% file: sections/sec_1_intro.tex
The task of (pairwise) \gls{AV} is to decide if two texts were written by the same person or not.
AV is traditionally performed by linguists who aim to uncover the authorship of anonymously written texts by inferring author-specific characteristics from the texts~\cite{BKA}.
Such characteristics are represented by so-called \textbf{linguistic features}.
They are derived from an analysis of errors (e.g. spelling mistakes), textual idiosyncrasies (e.g. grammatical inconsistencies) and stylistic patterns~\cite{BKA}.

Automated (machine-learning-based) systems have traditionally relied on so-called \textbf{stylometric features}~\cite{Stamatatos09}. Stylometric features tend to rely largely on linguistically motivated/inspired metrics.
The disadvantage of stylometric features is that their reliability is typically diminished when applied to texts with large topical variations.

Deep learning systems, on the other hand, can be developed to automatically learn \textbf{neural features} in an end-to-end manner~\cite{HRSN}. While these features can be learned in such a way that they are largely insensitive to the topic, on the negative side, they are generally not linguistically interpretable.


In this work we propose a substantial extension of our published \textsc{AdHominem} approach~\cite{DBLP:conf/bigdataconf/BoenninghoffHKN19}, in which we interpret the neural features produced by \textsc{AdHominem} not just from a metric point of view but, additionally, from a probabilistic point of view. 

With our modification of \textsc{AdHominem} we were also cognizant of the proposed future \gls{AV} shared tasks of the PAN organization~\cite{kestemont:2020}. Three broader research questions (cross-topic verification, open-set verification, and ``surprise task'') are put into the spotlight over the next three years.
In light of these challenges we define requirements for

\clearpage

\noindent automatically extracted neural features as follows:
%

\vspace{-0.05in}

\begin{itemize}
  \item \textbf{Distinctiveness:} Our extracted neural features should contain all necessary information w.r.t.~the \textit{writing style}, such that a verification system is able to distinguish same/different author/s in an open-set scenario. In order to automatically quantify deviations from the standard language, the text sample collection
  for the training phase must be sufficiently long.
  \item \textbf{Invariance:} Authors tend to shift the characteristics of their writing according to their situational disposition (e.g.~their emotional state)
  and the topic of the text/discourse.
  Extracted neural features should therefore, ideally, be invariant w.r.t.~the topic, the sentiment, the emotional state of the writer, and so forth. 
 \item \textbf{Robustness:} The writing style of a text can be influenced, for example, by a desire to imitate another author (e.g.~the original author of a fandom topic) or by applying a deliberate obfuscation strategy for other reasons. 
 Our extracted neural features should still lead to reliable verification results, even when obfuscation/imitation strategies are applied by the author. 
 \item \textbf{Adaptability:}
 The writing style is generally also affected by the type of the text, which is called genre. People change their linguistic register depending on the genre that they write in. This, in turn, leads to significant changes in the characteristics of the resulting text. For a technical system, it is thus extremely difficult to establish a common authorship between a WhatsApp message and a formal job application for example. In forensic disciplines, it is therefore important to train classifiers only on one genre at a time. In research, however, it is quite an interesting question how to, e.g., find a joint subspace representation/embedding for text samples across  different genres.
\end{itemize}

\vspace{-0.05in}

\noindent We assume that a single text sample has been written by a single person. If necessary, we need to examine a \textit{collaborative authorship} in advance~\cite{BKA}. Dealing with genre-adaption or obfuscation/imitation strategies is not part of the PAN 2020/21 \gls{AV} task. Another open question is the minimum size of a text sample required to obtain reliable output predictions. This question will also be left for future work.


%% file: sections/sec_2_theory.tex
Existing \gls{AV} algorithms can be taxonomically grouped w.r.t.~their design and characteristics, e.g.~instance- vs.~profile-based paradigms, intrinsic vs.~extrinsic methods~\cite{phdthesis}, or unary vs.~binary classification~\cite{HalvaniUnaryBinary}.
We may roughly describe the  work flow for a traditional binary AV classifier design as follows: In the feature engineering process, a set of manually defined stylometric features is extracted. Afterwards, a training and/or development set is used to fit a model to the data
and to tune possible hyper-parameters of the model. Typically, an additional calibration step is necessary to transform scores provided by the model into 
appropriate probability estimates.
Our modified \textsc{AdHominem} system works differently. We define a deep-learning model architecture with all of its hyper-parameters and thresholds a-priori and let the model learn suitable features for the provided setup on its own. As with most deep-learning approaches, the success of the proposed setup depends heavily on the availability of a large collection of text samples with many examples of representative variations in writing style.


The majority of published papers, using deep neural networks to build an \gls{AV} framework, have employed a classification loss~\cite{articlebagnall},~\cite{DBLP:conf/simbig/Litvak18}. However, metric learning objectives present a promising alternative~\cite{HRSN},~\cite{DBLP:journals/corr/abs-2003-11982}. The discriminative power of our proposed \gls{AV} method stems from a fusion of two well-known approaches into a single joint end-to-end learning procedure:
A precursor of our \textsc{AdHominem} system~\cite{DBLP:conf/bigdataconf/BoenninghoffHKN19} is used as a \textit{deep metric learning} framework~\cite{Hu14} 
to measure the similarity between two text samples. The features that are implicitly produced by the \textsc{AdHominem} system are then fed into a \gls{PLDA} layer~\cite{6466371} that functions as a pairwise discriminator to perform Bayes factor scoring in the learned metric space.

\vspace*{-0.15cm}
\subsection{Neural extraction of linguistic embedding vectors}
\vspace*{-0.15cm}
\label{seq:neural_ex}

\begin{figure*}[t]
    \centering
    \includegraphics[width=.9\textwidth]{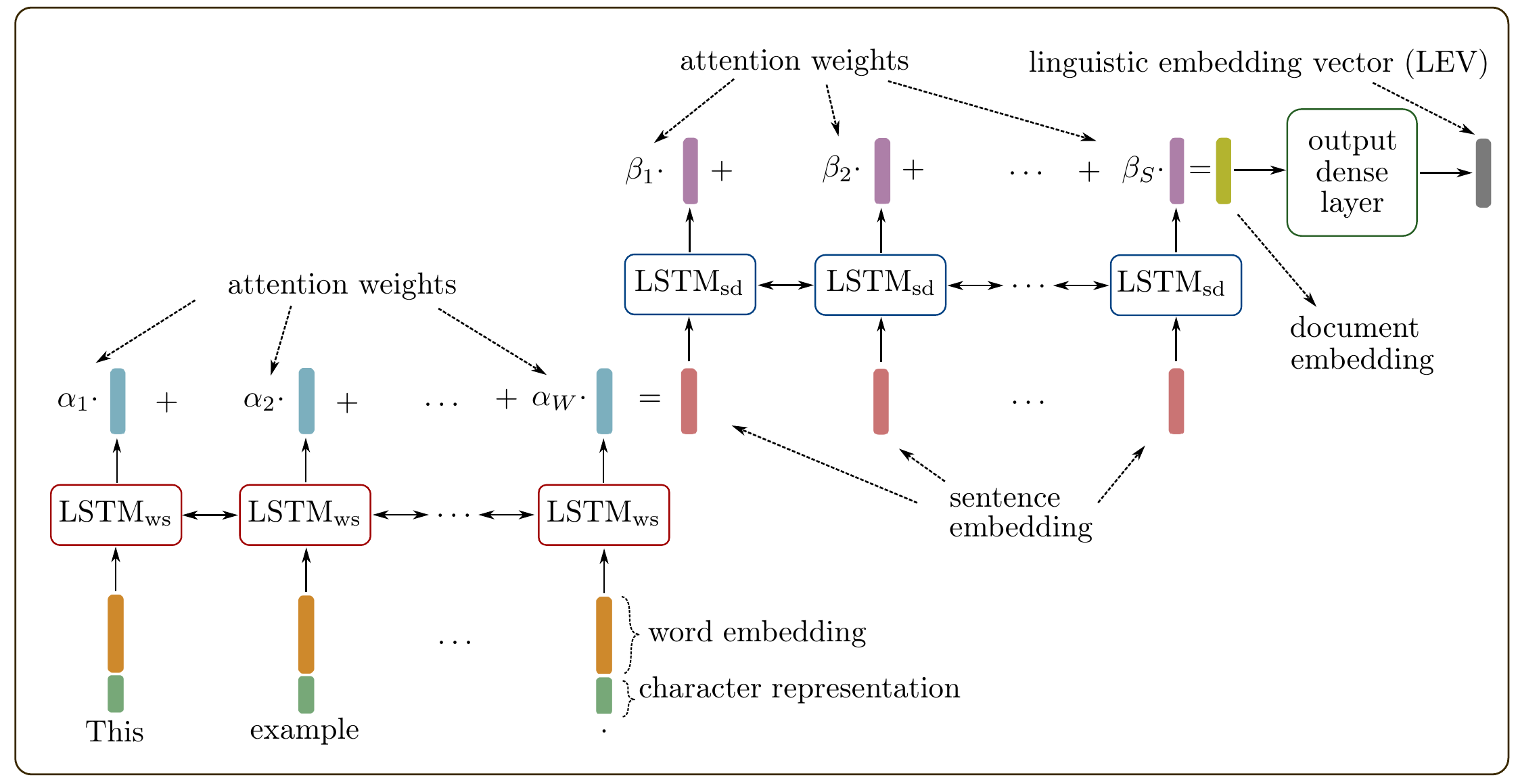}
    \vspace*{-0.1cm}
    \caption{Flowchart of the neural feature extraction as described in~\cite{DBLP:conf/bigdataconf/BoenninghoffHKN19}. A given text sample is transformed into the learned metric space by the function $\mathcal{A}_{\boldsymbol{\theta}}(\cdot)$.
    \vspace*{-.3cm}}
    \label{fig:AdHominem}
\end{figure*}

A text sample can be understood as a hierarchical structure of ordered discrete elements: It consists of a list of ordered sentences. Each sentence consists of an ordered list of tokens. Again, each token consists of an ordered list of characters.
The purpose of \textsc{AdHominem} is to map a document to a feature vector. 
More specifically, its Siamese topology includes a hierarchical neural feature extraction, which encodes the stylistic characteristics of a pair of documents
$(\mathcal{D}_1, \mathcal{D}_2)$, each of variable length,
into a pair of fixed-length \gls{LEVs} $\boldsymbol{y}_i$:
\begin{align}
 \boldsymbol{y}_i = \mathcal{A}_{\boldsymbol{\theta}}(\mathcal{D}_i) \in \mathbb{R}^{D\times 1}, ~i\in\{1,2\},
\end{align}
where $D$ denotes the dimension of the LEVs and $\boldsymbol{\theta}$ contains all trainable parameters. It is called a \textit{Siamese} network because both documents $\mathcal{D}_1$ and 
$\mathcal{D}_2$ are mapped through the exact same function~$\mathcal{A}_{\boldsymbol{\theta}}(\cdot)$. The internal structure of $\mathcal{A}_{\boldsymbol{\theta}}(\cdot)$ is illustrated in Fig.~\ref{fig:AdHominem}.
After preprocessing and tokenization (which will be explained in Section~\ref{sec:CATP}), the system passes a fusion of token and character embeddings into a two-tiered bidirectional LSTM~\cite{LSTM} network with attentions~\cite{Bahdanau14}. 
We incorporate a characters-to-word encoding layer to take the specific uses of prefixes and suffixes as well as spelling errors into account. An incorporation of attention layers allows us to visualize words and sentences that have been marked as ``highly  significant'' by the system. As shown in Fig.~\ref{fig:AdHominem}, the network produces document embeddings, which are converted into LEVs via a fully-connected dense layer. With this output layer, we can control the output dimension. \gls{AV} is accomplished by computing the Euclidean distance~\cite{Hu14} 
\begin{align}
    \label{eq:ED}
    d(\mathcal{D}_1, \mathcal{D}_2) 
        = \norm{\mathcal{A}_{\boldsymbol{\theta}}(\mathcal{D}_1) - \mathcal{A}_{\boldsymbol{\theta}}(\mathcal{D}_2)}^2
        = \norm{\boldsymbol{y}_1 - \boldsymbol{y}_2}^2
\end{align}
between both LEVs. If the distance in Eq.~\eqref{eq:ED} is above a given threshold $\tau$, then the system decides on \textit{different-authors}, if the distance is below $\tau$, then the system decides on \textit{same-authors}. 
Details are comprehensively described in~\cite{DBLP:conf/bigdataconf/BoenninghoffHKN19}.
\vspace*{-0.25cm}
\subsubsection{Pseudo-metric:}
\textsc{AdHominem} provides a framework to learn a pseuo-metric. Since we are using the Euclidean distance in Eq.~\eqref{eq:ED} we have the following properties:
\begin{align*}
  d(\mathcal{D}_1, \mathcal{D}_2) &\ge 0 \quad &\text{(nonnegativity)}\\
d(\mathcal{D}_1, \mathcal{D}_1) &= 0 \quad &\text{(identity)}\\
 d(\mathcal{D}_1, \mathcal{D}_2) &= d(\mathcal{D}_2, \mathcal{D}_1)  \quad &\text{(symmetry)}\\
 d(\mathcal{D}_1, \mathcal{D}_3) &\le d(\mathcal{D}_1, \mathcal{D}_2)
                                                    + d(\mathcal{D}_2, \mathcal{D}_3) \quad &\text{(triangle inequality)}
\end{align*}
Note that we may obtain  $d(\mathcal{D}_1, \mathcal{D}_2) = 0$ where $\mathcal{D}_1 \ne \mathcal{D}_2$.
\vspace*{-.3cm}
\subsubsection{Loss function:}
The entire network is trained end-to-end. 
For the pseudo-metric learning objective, we choose the modified contrastive loss~\cite{HRSN}:
\begin{align}
  \label{eq:loss1}
\loss_{\boldsymbol{\theta}} &= l \cdot \max \left\{ \norm{\boldsymbol{y}_1 -\boldsymbol{y}_2}^2 - \tau_s, 0 \right\}^2
 +       (1-l) \cdot \max \left\{ \tau_d - \norm{\boldsymbol{y}_1 -\boldsymbol{y}_2}^2 , 0\right\}^2,
\end{align}
where $l\in\{0,1\}$, $\tau_s<\tau_d$ and $\tau = \frac{1}{2}(\tau_s + \tau_d)$. During training, all distances between \textit{same-author} pairs are forced to stay below the lower of the two thresholds, $\tau_s$. Conversely, distances between \textit{different-authors} pairs are forced to remain above the higher threshold $\tau_d$.
By employing this dual threshold strategy, the system is made more insensitive to 
topical or intra-author variations between documents~\cite{Hu14},~\cite{DBLP:conf/bigdataconf/BoenninghoffHKN19}.
\vspace*{-0.15cm}
\subsection{Two-covariance model for Bayes factor scoring}
\vspace*{-0.15cm}
Text samples are characterized by a high variability. 
Statistical hypothesis tests can help to quantify the outputs/scores of our algorithm and 
to decide whether to accept or reject the decision. \textsc{AdHominem} can be extended with a framework for statistical hypothesis testing. More precisely, we are interested in the \gls{AV} problem where, given the LEVs of two documents, we have to decide for one of two hypotheses:
\begin{align*}
    \mathcal{H}_s:~~&\text{The two documents were written by the same person,} \\
    \mathcal{H}_d:~~&\text{The two documents were written by two different persons.}
\end{align*}

In the following, we will describe a particular case of the well-known \textit{probabilistic linear discriminant analysis} (PLDA)~\cite{10.1007/11744085_41}, which is also known as the \textit{two-covariance model}~\cite{6466371}.
Let us assume, the author's writing style is represented by a vector $\boldsymbol{x}$.  
We suppose that our (noisy) observed LEV $\boldsymbol{y} = \mathcal{A}_{\boldsymbol{\theta}}(\mathcal{D})$ stems from a Gaussian generative model that can be decomposed   as
\begin{align}
     \label{eq:LEV}
     \underbrace{\boldsymbol{y}}_{\text{linguistic embedding vector}} = \underbrace{\boldsymbol{x}}_{\text{author's writing style}} + 
     \underbrace{\boldsymbol{\epsilon}}_{\text{noise term}},
\end{align}
where $\boldsymbol{\epsilon}$ characterizes residual noise, caused by thematic varitions or by significant changes in the process of text production for instance. 

The idea behind this \textit{factor analysis} is that the writing characteristics of the author, measured in the observed LEV $\boldsymbol{y}$, lie in a latent variable $\boldsymbol{x}$.
The probability density functions for $\boldsymbol{x}$ and $\boldsymbol{\epsilon}$ in Eq.~\eqref{eq:LEV} are defined 
as in \cite{niko220}:
\begin{align}
    \label{eg:Gaussian1}
    p(\boldsymbol{x}) &= \mathcal{N}(\boldsymbol{x} | \boldsymbol{\mu}, \boldsymbol{B}^{-1}), \\
    \label{eg:Gaussian2}
    p(\boldsymbol{\epsilon}) &= \mathcal{N}(\boldsymbol{\epsilon} | \boldsymbol{0}, \boldsymbol{W}^{-1}),
\end{align}
where $\boldsymbol{B}^{-1}$ defines the \textit{between-author} covariance matrix and $\boldsymbol{W}^{-1}$ denotes the \textit{within-author} covariance matrix. As mentioned in~\cite{rummer1111}, the idea is to model \textit{inter-author variability} (with the covariance matrix $\boldsymbol{B}^{-1}$) and \textit{intra-author variability} (with the covariance matrix $\boldsymbol{W}^{-1}$).
From Eqs.~\eqref{eg:Gaussian1}$-$\eqref{eg:Gaussian2}, it can be deduced that the conditional density function is given by \cite{niko220}:
\begin{align}   
\label{eg:GaussianCon}
    p(\boldsymbol{y}|\boldsymbol{x})  &= \mathcal{N}(\boldsymbol{y} | \boldsymbol{x}, \boldsymbol{W}^{-1}).
\end{align}
%
Assuming we have a set of $n$ LEVs, $\mathcal{Y} = \{\boldsymbol{y}_1, \ldots \boldsymbol{y}_n \}$, verifiably associated to the same author, then we can compute the posterior (see Theorem 1 on page 175 in~\cite{GVK021834997}):
\begin{align}    
\label{eg:GaussianPos}
    p(\boldsymbol{x}|\mathcal{Y})  
        = \mathcal{N}(\boldsymbol{x} | \boldsymbol{L}^{-1}\boldsymbol{\gamma}, \boldsymbol{L}^{-1}),
\end{align}
where $\boldsymbol{L} = \boldsymbol{B}  + n \boldsymbol{W}$ 
and $\boldsymbol{\gamma} = \boldsymbol{B}\boldsymbol{\mu}   + \boldsymbol{W} \sum_{i=1}^{n} \boldsymbol{y}_i$.
Let us now consider the process of generating two \gls{LEV} 
$\boldsymbol{y}_i,~i\in \{1,2\}$. We 
have to distinguish between \textit{same-author} and \textit{different-author} pairs:

\subsubsection{Same-author pair:}
In the case of a same-author pair, a single latent vector $\boldsymbol{x}_0$ representing the author's writing style is generated from the prior $p(\boldsymbol{x})$ in Eq.~\eqref{eg:Gaussian1} and both LEVs 
$\boldsymbol{y}_i,~i\in \{1,2\}$ are generated from $p(\boldsymbol{y}|\boldsymbol{x}_0)$ in Eq.~\eqref{eg:GaussianCon}. The joint probability density function is then given by
\begin{align}
\label{eq:valid2}
 p(\boldsymbol{y}_1, \boldsymbol{y}_2 |\mathcal{H}_s) 
    = \frac{p(\boldsymbol{y}_1, \boldsymbol{y}_2| 
               ~ \boldsymbol{x}_0, \mathcal{H}_s) 
                ~p(\boldsymbol{x}_0|\mathcal{H}_s)}
            {p(\boldsymbol{x}_0| \boldsymbol{y}_1, \boldsymbol{y}_2, \mathcal{H}_s)} 
   = \frac{p(\boldsymbol{y}_1|\boldsymbol{x}_0)
           ~ p(\boldsymbol{y}_2|\boldsymbol{x}_0)
                ~p(\boldsymbol{x}_0)}
             {p(\boldsymbol{x}_0| \boldsymbol{y}_1, \boldsymbol{y}_2)}.
\end{align}
The term $p(\boldsymbol{x}_0| \boldsymbol{y}_1, \boldsymbol{y}_2)$ can be computed using Eq.~\eqref{eg:GaussianPos}.

\subsubsection{Different-authors pair:}
For a different-authors pair, two latent vectors, $\boldsymbol{x}_i$ for $i\in \{1,2\}$, representing two different authors' writing characteristics, are independently generated from $p(\boldsymbol{x})$ in Eq.~\eqref{eg:Gaussian1}. The corresponding LEVs $\boldsymbol{y}_i$ are generated from $p(\boldsymbol{y}|\boldsymbol{x}_{i})$ in Eq.~\eqref{eg:GaussianCon}. The joint probability density function is then given by
\begin{align}
\label{ref:valid1}
 p(\boldsymbol{y}_1, \boldsymbol{y}_2 |\mathcal{H}_d) 
    = p(\boldsymbol{y}_1 |\mathcal{H}_d) 
           ~ p(\boldsymbol{y}_2 |\mathcal{H}_d) 
    = \frac{p(\boldsymbol{y}_1|\boldsymbol{x}_1)
            p(\boldsymbol{x}_1)    
            }{p(\boldsymbol{x}_1|\boldsymbol{y}_1)}
            \cdot
            \frac{p(\boldsymbol{y}_2|\boldsymbol{x}_2)
            p(\boldsymbol{x}_2)    
            }{p(\boldsymbol{x}_2|\boldsymbol{y}_2)}.
\end{align}
The terms $p(\boldsymbol{x}_1|\boldsymbol{y}_1)$ and $p(\boldsymbol{x}_2|\boldsymbol{y}_2)$ are again obtained from Eq.~\eqref{eg:GaussianPos}.

\subsubsection{Verification score:}
The described probabilistic model involves two steps: a training phase to learn the parameters of the Gaussian distributions in Eqs.~\eqref{eg:Gaussian1}$-$\eqref{eg:Gaussian2} and a verification phase to infer whether both text samples come from the same author.

\noindent
For both steps, we need to define the verification score, which can now be calculated as the log-likelihood ratio between the two hypotheses $\mathcal{H}_s$ and $\mathcal{H}_d$:
\begin{align}
\label{eq:verifscore}
 \text{score}(\boldsymbol{y}_1, \boldsymbol{y}_2) 
 &= \log p(\boldsymbol{y}_1, \boldsymbol{y}_2 |\mathcal{H}_s) - \log p(\boldsymbol{y}_1, \boldsymbol{y}_2 |\mathcal{H}_d) 
 \nonumber \\
 &=        \log p(\boldsymbol{x}_0)
         - \log p(\boldsymbol{x}_1) 
         - \log p(\boldsymbol{x}_2)
    \nonumber  \\ &\quad
            +  \log p(\boldsymbol{y}_1|\boldsymbol{x}_0)  
            + \log p(\boldsymbol{y}_2|\boldsymbol{x}_0)
            - \log p(\boldsymbol{y}_1|\boldsymbol{x}_1) 
            - \log p(\boldsymbol{y}_2|\boldsymbol{x}_2)
    \nonumber  \\ &\quad
         - \log p(\boldsymbol{x}_0| \boldsymbol{y}_1, \boldsymbol{y}_2) 
            + \log p(\boldsymbol{x}_1|\boldsymbol{y}_1) 
            + \log p(\boldsymbol{x}_2|\boldsymbol{y}_2) 
\end{align}
Eq.~\eqref{eq:verifscore} is often called the \textit{Bayes factor}.
Since $p(\boldsymbol{y}_1, \boldsymbol{y}_2 |\mathcal{H}_s)$ in Eq.~\eqref{eq:valid2} and $p(\boldsymbol{y}_1, \boldsymbol{y}_2 |\mathcal{H}_d)$ in Eq.~\eqref{ref:valid1} are independent of $\boldsymbol{x}_0$ and $\boldsymbol{x}_1$, $\boldsymbol{x}_2$, we can choose any values for the latent variables, as long as the denominator is non-zero~\cite{niko220}.
Substituting Eqs.~\eqref{eg:Gaussian1},~\eqref{eg:Gaussian2},~\eqref{eg:GaussianCon},~\eqref{eg:GaussianPos} in Eq.~\eqref{eq:verifscore} and selecting $\boldsymbol{x}_0 = \boldsymbol{x}_1 = \boldsymbol{x}_2 = \boldsymbol{0}$, we obtain~\cite{6466371}
\begin{align}
\label{eq:verifscore2}
 \text{score}(\boldsymbol{y}_1, \boldsymbol{y}_2) 
 &=       - \log \mathcal{N}(\boldsymbol{0} | \boldsymbol{\mu}, \boldsymbol{B}^{-1}) 
            - \log  \mathcal{N}(\boldsymbol{0} | \boldsymbol{L}_{1,2}^{-1}\boldsymbol{\gamma}_{1,2}, \boldsymbol{L}_{1,2}^{-1})
        \nonumber  \\ &\quad
            + \log  \mathcal{N}(\boldsymbol{0} | \boldsymbol{L}_{1}^{-1}\boldsymbol{\gamma}_{1}, \boldsymbol{L}_{1}^{-1})
            + \log  \mathcal{N}(\boldsymbol{0} | \boldsymbol{L}_{2}^{-1}\boldsymbol{\gamma}_{2}, \boldsymbol{L}_{2}^{-1}), 
    \\[-1em] \nonumber
\end{align}
where $\boldsymbol{L}_{1,2} = \boldsymbol{B}  + 2 \boldsymbol{W}$, $\boldsymbol{\gamma}_{1,2} = \boldsymbol{B}\boldsymbol{\mu}   + \boldsymbol{W} (\boldsymbol{y}_1 + \boldsymbol{y}_2)$ and
$\boldsymbol{L}_{i} = \boldsymbol{B}  + \boldsymbol{W}$, $\boldsymbol{\gamma}_{i} = \boldsymbol{B}\boldsymbol{\mu}   + \boldsymbol{W} \boldsymbol{y}_i$ for $i\in \{1,2\}$. As described in~\cite{niko220}, the score in Eq.~\eqref{eq:verifscore2} can now be rewritten as
\begin{align}
\nonumber\\[-1em]
 \text{score}(\boldsymbol{y}_1, \boldsymbol{y}_2) = \boldsymbol{y}_1 \boldsymbol{\Lambda} \boldsymbol{y}_2^T
   + \boldsymbol{y}_2 \boldsymbol{\Lambda} \boldsymbol{y}_1^T
   + \boldsymbol{y}_1 \boldsymbol{\Gamma} \boldsymbol{y}_1^T
   + \boldsymbol{y}_2 \boldsymbol{\Gamma} \boldsymbol{y}_2^T
   + \big(\boldsymbol{y}_1 + \boldsymbol{y}_2 \big)^T \boldsymbol{\rho}
   + \kappa,
   \label{eq:scorequadratic}
\\[-1em] \nonumber
\end{align}
where the parameters $\boldsymbol{\Lambda}$, $\boldsymbol{\Gamma}$, $\boldsymbol{\rho}$ and $\kappa$ 
of the quadratic function in Eq.~\eqref{eq:scorequadratic} are given by
\begin{align*}
 \boldsymbol{\Gamma} 
    &= \frac{1}{2} \boldsymbol{W}^T
            \big(\boldsymbol{\widetilde{\Lambda}}
            - \boldsymbol{\widetilde{\Gamma}} \big)
            \boldsymbol{W} ,
& \boldsymbol{\Lambda} 
     &= \frac{1}{2} \boldsymbol{W}^T \boldsymbol{\widetilde{\Lambda}}
    \boldsymbol{W},
\\ \boldsymbol{\rho} 
   & =  \boldsymbol{W}^T
            \big(\boldsymbol{\widetilde{\Lambda}}
            - \boldsymbol{\widetilde{\Gamma}} \big)
            \boldsymbol{B} \boldsymbol{\mu},
&  \kappa 
  & = \widetilde{\kappa} + \frac{1}{2}
     \bigg( \big(\boldsymbol{B} \boldsymbol{\mu} \big)^T 
     \big(\boldsymbol{\widetilde{\Lambda}}
            - 2\boldsymbol{\widetilde{\Gamma}} \big)
     \boldsymbol{B} \boldsymbol{\mu} \bigg)
\end{align*}
and the auxiliary variables are
\begin{align*}
 \boldsymbol{\widetilde{\Gamma}}
    &= \big(\boldsymbol{B} + \boldsymbol{W}
        \big)^{-1} ,
 \quad \quad \boldsymbol{\widetilde{\Lambda}}
    = \big(\boldsymbol{B} +2  \boldsymbol{W}
        \big)^{-1},
\\
\widetilde{\kappa}
   & =2 \log \det \big( \boldsymbol{\widetilde{\Gamma}} \big)
   -  \log \det \big(\boldsymbol{B} \big)
  -  \log \det \big(\boldsymbol{\widetilde{\Lambda}}\big)
   + \boldsymbol{\mu}^T  \boldsymbol{B}  \boldsymbol{\mu} .
\end{align*}
Hence, the verification score is a symmetric quadratic function of the LEVs. 
The  probability for a same-author trial can be computed from the log-likelihood ratio score as follows:

\vspace{-0.75cm}

\begin{align}
 \label{eq:llrs}
 p(\mathcal{H}_s|\boldsymbol{y}_1, \boldsymbol{y}_2)
        &= \frac{p(\mathcal{H}_s) ~ p(\boldsymbol{y}_1, \boldsymbol{y}_2|\mathcal{H}_s)}
            {p(\mathcal{H}_s) ~ p(\boldsymbol{y}_1, \boldsymbol{y}_2|\mathcal{H}_s) + p(\mathcal{H}_d) ~ p(\boldsymbol{y}_1, \boldsymbol{y}_2|\mathcal{H}_d)}
\end{align}
The \gls{AV} datasets provided by the PAN organizers are balanced w.r.t. authorship labels. Hence, we can assume $p(\mathcal{H}_s) = p(\mathcal{H}_d) = \frac{1}{2}$.
We can rewrite Eq.~\eqref{eq:llrs} as follows:
\begin{align}
 \label{eq:scorefinal} 
  p(\mathcal{H}_s|\boldsymbol{y}_1, \boldsymbol{y}_2)
        &= \frac{p(\boldsymbol{y}_1, \boldsymbol{y}_2|\mathcal{H}_s)}
            {p(\boldsymbol{y}_1, \boldsymbol{y}_2|\mathcal{H}_s) + p(\boldsymbol{y}_1, \boldsymbol{y}_2|\mathcal{H}_d)}
= \text{Sigmoid}\big( \text{score}(\boldsymbol{y}_1, \boldsymbol{y}_2) \big)
\end{align}

\vspace*{-0.25cm}

\subsubsection{Loss function:}
To learn the probabilistic layer, we incorporate Eq.~\eqref{eq:scorefinal} into the binary cross entropy:
\begin{align}
\label{eq:loss2}
\loss_{\boldsymbol{\phi}} 
    = l \cdot \log \left\{ p(\mathcal{H}_s|\boldsymbol{y}_1, \boldsymbol{y}_2) \right\}
            + (1-l) \cdot \log \left\{ 1 - p(\mathcal{H}_s|\boldsymbol{y}_1, \boldsymbol{y}_2) \right\},
\end{align}
where $\boldsymbol{\phi} = \big\{ \boldsymbol{W}, \boldsymbol{B}, \boldsymbol{\mu} \big\}$ contains the trainable parameters
of the probabilistic layer.
\vspace*{-0.25cm}
\subsubsection{Cholesky decomposition for numerically stable covariance training:}
We can treat $\boldsymbol{\phi} = \big\{ \boldsymbol{W}, \boldsymbol{B}, \boldsymbol{\mu} \big\}$ given by Eqs.~\eqref{eg:Gaussian1}$-$\eqref{eg:Gaussian2} as trainable paramters in our deep learning framework. 
For both covariance matrices we need to guarantee positive definiteness. 
Instead of learning $\boldsymbol{W}$ and $\boldsymbol{B}$ directly, we enforce the positive definiteness of them through Cholesky decomposition 
by constructing trainable lower-triangular matrices $\boldsymbol{L}_{W}$ and $\boldsymbol{L}_{B}$ with exponentiated (positive) diagonal elements. The 
estimated covariance matrices are constructed via $\boldsymbol{\widehat{W}} = \boldsymbol{L}_{W} ~\boldsymbol{L}_{W}^T$ and $\boldsymbol{\widehat{B}} = \boldsymbol{L}_{B} ~\boldsymbol{L}_{B}^T$.
We computed and updated the gradients of $\boldsymbol{L}_{W}$ and $\boldsymbol{L}_{B}$ with respect to the loss function in Eq.~\eqref{eq:loss2}. 
\vspace*{-0.25cm}
\subsection{Ensemble inference}
Neural networks are randomly initialized, trained on the same, but shuffled data and affected by regularization techniques like dropout. Hence, they will find a different set of weights/biases each time, which in turn produces different predictions. 
To reduce the variance, we propose to train an ensemble of models and to combine the predictions of 
Eq.~\eqref{eq:scorefinal} from these models,
\begin{align}
\label{eq:ensenble}
    \mathbb{E}\big[ p(\mathcal{H}_s|\boldsymbol{y}_1, \boldsymbol{y}_2)
    \big]
    \approx \frac{1}{m}\sum_{i=1}^m p_{\mathcal{M}_i}(\mathcal{H}_s|\boldsymbol{y}_1, \boldsymbol{y}_2),
\end{align}
where $\mathcal{M}_i$ indicates the $i$-th trained model. 
Finally, we determine the non-answers for predicted probabilities, i.e.  
$\mathbb{E}\big[ p(\mathcal{H}_s|\boldsymbol{y}_1, \boldsymbol{y}_2)\big] = 0.5$,
if $0.5 -\delta  <  \mathbb{E}\big[ p(\mathcal{H}_s|\boldsymbol{y}_1, \boldsymbol{y}_2)\big] < 0.5 + \delta$. Parameter $\delta$ can be found by applying a simple grid search.


%% file: sections/sec_3_implementation.tex
\label{sec:CATP}
The 2020 edition of the PAN authorship verification task focuses on \textit{fanfiction} texts, fictional texts written by fans of previous, original literary works that have become popular like "Harry Potter".
Usually, authors of fanfiction 
preserve core elements of the storyline by reusing main characters and settings. Nevertheless, they may also contain changes or alternative interpretations of some parts of the known storyline.   
The subject area of the original work is called \textit{fandom}. 
The PAN organizers are providing unique author and fandom (topical) labels for all fanfiction pairs. 
The dataset has been derived from the corpus compiled in~\cite{bischoff:2020}.
A detailed description of the dataset is given in~\cite{kestemont:2020}.  

As mentioned in the introduction, automatically extracted neural features should be invariant w.r.t.~shifts in topic and/or sentiment. Ideally, LEVs should only contain information regarding the writing style of the authors. What is well-established in automatic AV is that the topic of a text generally matters. What is still not clear, however, is how stylometric or neural features are influenced/affected by the topic (i.e.~fandom in this case).
To increase the generalization capabilities of our model and to increase the model's resilience towards cross-topic fanfiction pairs we devised the following preprocessing strategies, as outlined in Sections 3.1 through 3.3.
\vspace*{-0.4cm}
\subsection{Topic masking}
\vspace*{-0.1cm}
Experiments show that considering a large set of token types can lead to significant overfitting effects.
To overcome this, we reduced the vocabulary size for tokens as well as for characters by mapping all rare token/character types to a special \textit{unknown} (\texttt{<UNK>}) token. 
This is quite similar to the text distortion approach proposed in~\cite{stamatatos-2017-authorship}. However, even when a rare/misspelled token is replaced by the \texttt{<UNK>} token, it can still be encoded by the character representation.
%
%
%
\begin{figure}[t]
 \centerline{\includegraphics[width=.95\textwidth]{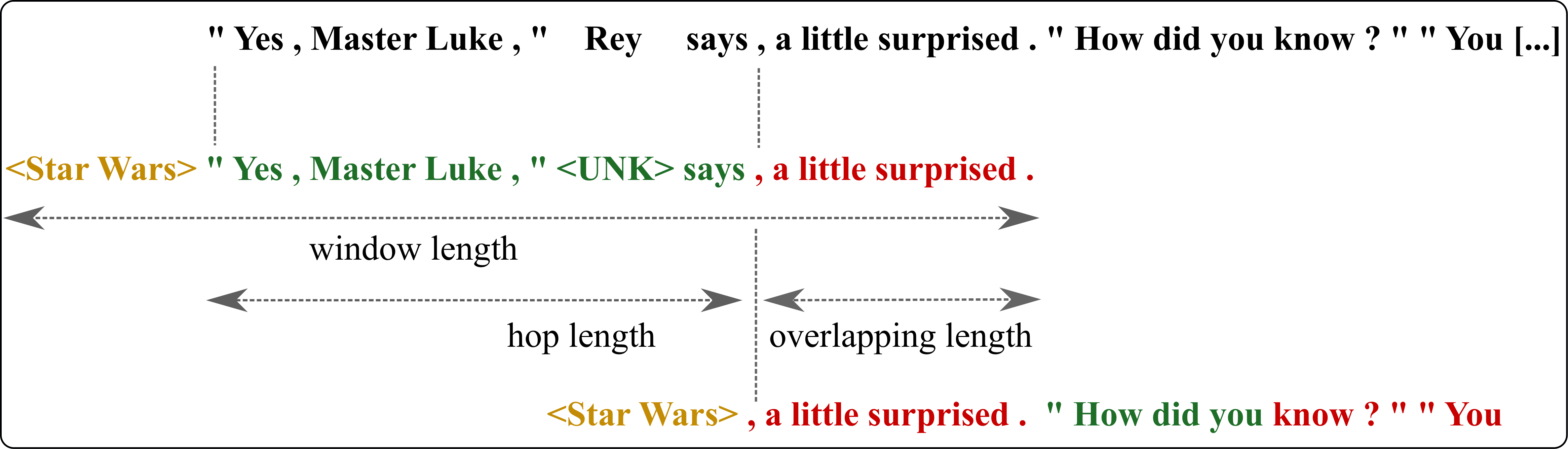}}
 \vspace*{-0.1cm}
 \caption{Example of our sliding window approach with contextual prefix. 
 \vspace*{-1.cm}
 }
 \label{fig:sli_wi}
\end{figure}
\vspace*{-0.4cm}
\subsection{Sliding window with contextual prefix}
\vspace*{-0.1cm}
Fanfiction frequently contains dialogues and quoted text. Sentence boundary detectors, therefore, tend to be very error prone and steadily fail to segment the data into appropriate sentence units. 
We decided to perform tokenization without strict sentence boundary detection and generated \textit{sentence-like units} via a sliding window technique instead. An example that illustrates the procedure is shown in Fig.~\ref{fig:sli_wi}. We used overlapping windows to guarantee that semantically and grammatically linked neighboring tokens are located in the same unit. 
We also added a \textit{contextual prefix} which is provided by the fandom labels. To initialize the prefix embeddings, we removed all non-ASCII characters, tokenized the fandom string and averaged the corresponding word embeddings. The final sliding window length (in tokens) is given by {\color{rub_green}hop\_length} + {\color{rub_red}overlapping\_length} + {\color{rub_orange}1}.
\begin{figure}[t]
\hspace*{-.85cm}
\begin{minipage}[t]{.62\textwidth}
\centering
\scalebox{0.8}{
\begin{algorithm}[H]
	\DontPrintSemicolon
	\small
	\textbf{Input: }$\mathcal{A}^{(1)}, \ldots \mathcal{A}^{(N)} $\;
	\textbf{Output: }$\mathcal{G}^{(1)}, \mathcal{G}^{(2)}$
	\BlankLine
    Initialize $\mathcal{G}^{(i)} = \{\emptyset \}$ for $i\in \{1,2,3\}$ \;
	\For{$i=1, \ldots, N$}{
		\uIf{$\vert \mathcal{A}^{(i)}\vert = \vert\{(a^{(i)}, f^{(i)}, d^{(i)}) \}\vert = 1$}{
			\tcp{Authors with single doc}
			$\mathcal{G}^{(1)} \longleftarrow \mathcal{G}^{(1)} \cup \{  (a^{(i)}, f^{(i)}, d^{(i)})\}$
		}
		\uElseIf{$\vert \mathcal{A}^{(i)}\vert > 1$ {\bf and} $\vert \mathcal{A}^{(i)}\vert \bmod 2 = 0$}{
           \tcp{Number of docs = 2,4,..}
			$\mathcal{G}^{(2)} \longleftarrow \mathcal{G}^{(2)} \cup \{   \mathcal{A}^{(i)}\}$
		}
		\ElseIf{$\vert \mathcal{A}^{(i)}\vert > 1$ {\bf and} $\vert \mathcal{A}^{(i)}\vert \bmod 2 = 1$}{
		\tcp{Number of  docs = 3,5,..}
			$\mathcal{G}^{(3)} \longleftarrow \mathcal{G}^{(3)} \cup \{ \mathcal{A}^{(i)} \}$
		}
	}
	\tcc{Assign one doc of authors in group 3 to group 1, assign all remaining docs to group 2.}
	\ForAll{$\mathcal{A} \in \mathcal{G}^{(3)}$}{
		randomly draw $(a, f, d) \in \mathcal{A}$ \;
		$\mathcal{G}^{(1)} \longleftarrow \mathcal{G}^{(1)} \cup \{ (a, f, d)\}$ \;
        $\mathcal{G}^{(2)} \longleftarrow \mathcal{G}^{(2)} \cup \{ \mathcal{A} \setminus \{(a, f, d)\} \}$ \;
	}
\caption{\textsc{MakeTwoGroups}\label{al1}}
\end{algorithm}
}
\end{minipage}\medskip
\hspace*{-1.5cm}
\begin{minipage}[t!]{.62\textwidth}
\centering
\vspace*{-1.052cm}
\scalebox{0.8}{
\begin{algorithm}[H]
	\DontPrintSemicolon
	\small
\textbf{Input: }$\mathcal{A}, \mathcal{D}, \mathcal{G}^{(1)}, \mathcal{G}^{(2)}$ \;
	\textbf{Output: }$\mathcal{D}, \mathcal{G}^{(1)}, \mathcal{G}^{(2)}$
	\BlankLine
        \BlankLine
		\uIf{$|\mathcal{A}| > 1$}{
				\tcc{Add to group 2, if at least two docs remain.}
			$\mathcal{G}^{(2)} \longleftarrow \mathcal{G}^{(2)} \cup \{\mathcal{A}\}$	
            }
    \ElseIf{$ \vert \mathcal{A} \vert = \vert \{(a, f, d)\} \vert = 1$}{
           \tcc{Add to group 1, if only one doc remains and no doc was assigned to group 1 via MakeTwoGroups(). 
        Otherwise, make another same-author pair.}
           \uIf{$\forall(a', f', d') \in  \mathcal{G}^{(1)} ~ \exists ~a'~:~ a = a'$}{
                        $\mathcal{D} \longleftarrow \mathcal{D} \cup \{(d, d', f, f', 1)\}$\;
                        $\mathcal{G}^{(1)} \longleftarrow \mathcal{G}^{(1)} 
                                        \setminus \{(a', f', d') \}$ \;
                    }
                    \Else{
                        $\mathcal{G}^{(1)} \longleftarrow \mathcal{G}^{(1)} \cup \{ (a, f, d)\}$}
        }
\caption{\textsc{CleanAfterSampling}\label{al2}}
\end{algorithm}
}
\end{minipage}
\vspace*{-0.8cm}
\end{figure}

\vspace*{-0.4cm}
\subsection{Data split and augmentation}
\vspace*{-0.1cm}

To tune our model we split the datasets into a \textit{train} and a \textit{dev} set. Table~\ref{tab:split} shows the resulting sizes.
\begin{table}[b]
    \vspace*{-0.3cm}
    \centering
      \begin{tabular}{ r | r r r}
                   & \textbf{train set~~}    & \textbf{dev set~~}    
                   &\textbf{ test set~~~}\\ \hline
    small dataset~ &~47,340 pairs~ &~5,261 pairs~  &~\multirow{2}{*}{14,311 pairs} \\
   large dataset~ &~261,786 pairs~ &~13,779 pairs~  &
  \end{tabular}
    \caption{Dataset sizes (including the provided test set) after splitting.}
    \label{tab:split}
\end{table}
The size of the train set can then be increased synthetically by dissembling all predefined document pairs and re-sampling new same-author and different-author pairs in each epoch. 
We first removed all documents in the {\em train\/} set which also appear in the {\em dev\/} set. Afterwards, we reorganized the {\em train\/} set as described in Alg.~\ref{al1}~-~\ref{al3}.
Assuming the $i$-th author with $i\in\{1,\ldots,N\}$ contributes with $N_i$ fanfiction texts,
we define a set $\mathcal{A}^{(i)} = \{(a^{(i)}, f_1^{(i)} , d^{(i)}_1), \ldots, (a^{(i)}, f_{N_i}^{(i)},  d^{(i)}_{N_i}) \}$ containing $3$-tuples of the form $(a^{(i)}, d_{j}^{(i)}, f_{j}^{(i)})$, where $a^{(i)}$ is the author ID, $d_{j}^{(i)}$ represents the $j$-th document and $f_{j}^{(i)}$ is the corresponding fandom label. The objective is to obtain a new set $\mathcal{D}$ of re-sampled pairs, containing $5$-tuples of the form $(d^{(1)}, d^{(2)}, f^{(1)}, f^{(2)}, l)$, where $d^{(1)}$, $d^{(2)}$ defines the sampled fanfiction pair, $f^{(1)}$, $f^{(2)}$ are the  corresponding fandom labels and $l \in \{0,1\}$ indicates whether the texts are written by the same author ($l=1$) or by different authors ($l=0$).
We obtain re-sampled pairs via $\mathcal{D}=\textsc{SamplePairs}(\mathcal{A}^{(1)}, \ldots, \mathcal{A}^{(N)})$ in Alg.~\ref{al3}. The epoch-wise sampling of new pairs can be accomplished beforehand to speed up the training phase.

\begin{figure}[t]
\centering
\scalebox{0.8}{
\begin{algorithm}[H]
	\DontPrintSemicolon
    \small
	\textbf{Input: }$\mathcal{A}^{(i)} = \{(a^{(i)}, f_1^{(i)} , d^{(i)}_1), \ldots, (a^{(i)}, f_{N_i}^{(i)},  d^{(i)}_{N_i}) \}~ \forall i\in\{1, \ldots N\}$ \;
	\textbf{Output: }$\mathcal{D}$
	\BlankLine
	Initialize $\mathcal{D} = \{\emptyset \}$ \;
	$\{\mathcal{G}^{(1)}, \mathcal{G}^{(2)} \} = \textsc{MakeTwoGroups}
            (\mathcal{A}^{(1)}, \ldots \mathcal{A}^{(N)})$\;
	\While{$\vert \mathcal{G}^{(2)}\vert > 0$ {\bf or} $\vert \mathcal{G}^{(1)}\vert > 1 $}{
	\tcp{Sample same-author pair}
	\If{$\vert \mathcal{G}^{(2)}\vert > 0$}{
            randomly draw $\mathcal{A} \in \mathcal{G}^{(2)}$ \;
            $\mathcal{G}^{(2)} \longleftarrow \mathcal{G}^{(2)} \setminus \{\mathcal{A}\}$ \;
            randomly draw $(a, f_1, d_1), (a,f_2, d_2) \in \mathcal{A}$ \;
			$\mathcal{A} \longleftarrow \mathcal{A} \setminus \{(a,f_1, d_1),(a,f_2, d_2) \}$ \;
			$\mathcal{D} \longleftarrow \mathcal{D} \cup \{(d_1, d_2, f_1, f_2, 1)\}$\\
			$\{\mathcal{D}, \mathcal{G}^{(1)}, \mathcal{G}^{(2)}\} \longleftarrow$ 
			\textsc{CleanAfterSampling}$(\mathcal{A}, \mathcal{D}, \mathcal{G}^{(1)}, \mathcal{G}^{(2)})$ \;
    }
    \tcp{Sample different-authors pair}
	\uIf{$\vert \mathcal{G}^{(1)}\vert >1$}{ 
			randomly draw $(a^{(1)}, f^{(1)}, d^{(1)}) \in \mathcal{G}^{(1)}$ and $(a^{(2)}, f^{(2)}, d^{(2)}) \in \mathcal{G}^{(1)}$ \;
			$\mathcal{G}^{(1)} \longleftarrow \mathcal{G}^{(1)} \setminus \{(a^{(1)}, f^{(1)}, d^{(1)}), (a^{(2)}, f^{(2)}, d^{(2)})\}$ \;
            $\mathcal{D} \longleftarrow \mathcal{D} \cup \{(d^{(1)}, d^{(2)}, f^{(1)}, f^{(2)}, 0)\}$\\
		}
    \ElseIf{$\vert \mathcal{G}^{(2)}\vert > 1$}{
            randomly draw $\mathcal{A}^{(1)},\mathcal{A}^{(2)}  \in \mathcal{G}^{(2)}$\;
            $\mathcal{G}^{(2)} \longleftarrow \mathcal{G}^{(2)} 
            \setminus \{\mathcal{A}^{(1)} \mathcal{A}^{(2)}\}$ \;
            randomly draw $(a^{(1)}, f^{(1)}, d^{(1)}) \in \mathcal{A}^{(1)}$ 
                    and $(a^{(2)}, f^{(2)}, d^{(2)}) \in \mathcal{A}^{(2)}$ \;
			$\mathcal{A}^{(1)} \longleftarrow \mathcal{A}^{(1)} \setminus \{(a^{(1)},f^{(1)}, d^{(1)})\}$ 
                and $\mathcal{A}^{(2)} \longleftarrow \mathcal{A}^{(2)} 
                            \setminus \{(a^{(2)},f^{(2)}, d^{(2)})\}$ \;
            $\mathcal{D} \longleftarrow \mathcal{D} \cup \{(d^{(1)}, d^{(2)},f^{(1)},  f^{(2)}, 0)\}$\;
            \For{$\mathcal{A} \in \{\mathcal{A}^{(1)},\mathcal{A}^{(2)}\}$}
            {$\{\mathcal{D}, \mathcal{G}^{(1)}, \mathcal{G}^{(2)}\} \longleftarrow$ 
			\textsc{CleanAfterSampling}$(\mathcal{A}, \mathcal{D}, \mathcal{G}^{(1)}, \mathcal{G}^{(2)})$ 
            }
    }
    }
\caption{\textsc{SamplePairs}\label{al3}}
\end{algorithm}
}
\vspace*{-.8cm}
\end{figure}

%% file: sections/sec_4_results.tex
\begin{figure}[t]
\begin{floatrow}
\hspace*{-1.45cm}
\ffigbox{
    \includegraphics[width=.3\textwidth]{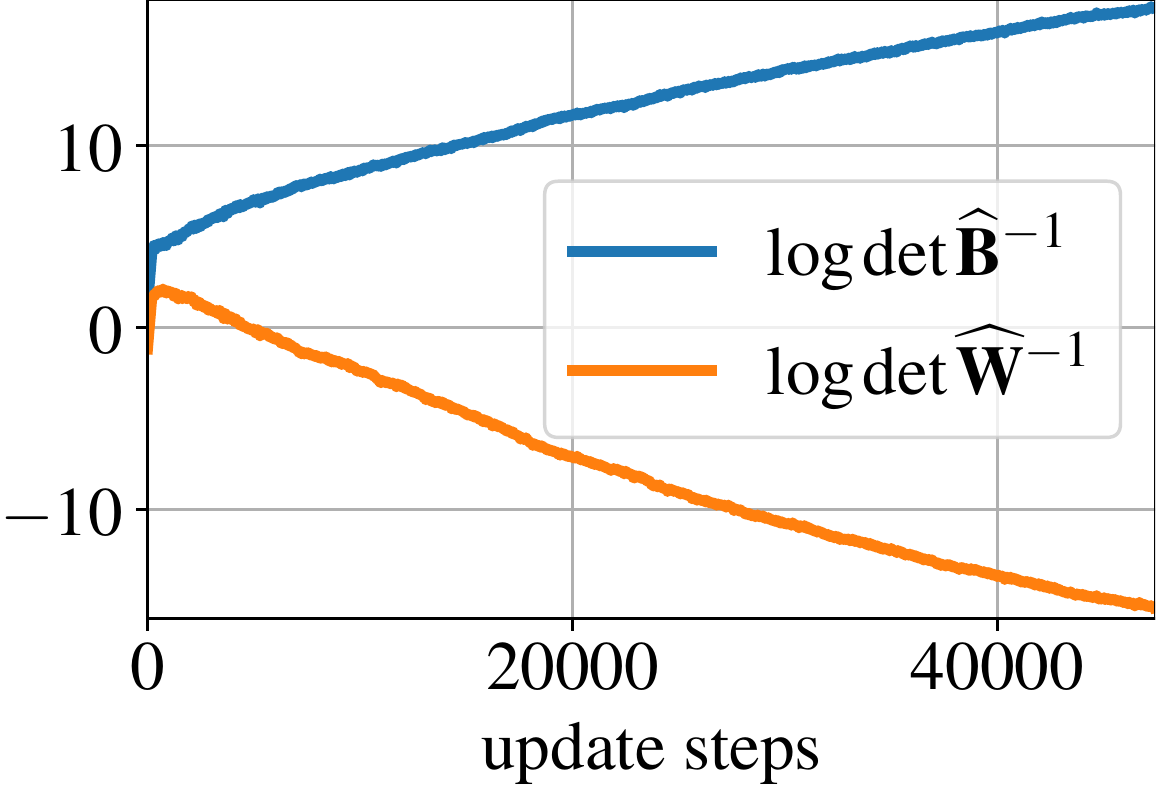}
    }{
    \caption{Entropy curves.}
    \label{fig:en_cur}
    }
\hspace*{-1.75cm}
\capbtabbox{
    \small
    \scalebox{0.8}{
    \begin{tabular}{r c | c  c |c c c c c}
    &  \textsc{AdHominem}
    &~train set~ &~evaluation  &\texttt{~AUC~}           &\texttt{~c@1~}       &\texttt{f\_05\_u}    &\texttt{~F1~}        &\texttt{overall} 
    \\ \hline
    1~&~    early-bird~ &~small~  &~dev set~ &~0.964~  &~0.919~  &~0.916~  &~0.932~  &~0.933~ \\
    2~&~   ensemble~   &~small~  &~dev set~      &~0.977~  &~0.942~    &~0.938~  &~0.946~  &~0.951~ \\
    3~&~ensemble~ &~large~  &~dev set~ &~0.985~  &~0.955~  &~0.940~  &~0.959~  &~0.960~
    \\ \hline
    4~&~    single~     &~small~  &~dev set~      &~0.975~  &~0.943~     &~0.921~  &~0.951~  &~0.948~ \\
    5~&~single~ & ~ large~    &~dev set~ &~0.983~  &~0.950~  &~0.944~  &~0.954~  &~0.958~ 
    \\\hline
    6~&~ early-bird~ &~small~ &~test set~    &~0.923~  &~0.861~  &~0.857~  &~0.891~  &~0.883~ \\
    7~&~ ensemble~ &~small~   &~test set~     &~0.940~  &~0.889~  &~0.853~  &~0.906~  &~0.897~ \\
    8~&~ ensemble~ &~large~    &~test set~   &~0.969~  &~0.928~  &~0.907~  &~0.936~  &~0.935~
    \end{tabular}
    }
    }{
    \caption{Results w.r.t the provided metrics on the dev and test sets.}
    \label{tab:res}
    }
\end{floatrow}
\end{figure}

Table~\ref{tab:res} reports the evaluation results\footnote{\small The source code will be publicly available to interested readers after the peer review notification, including the set of hyper-parameters.} for our proposed system over the {\em dev\/} set and the {\em test\/} set\footnote{The {\em test\/} set was not accessible to the authors. Results on the {\em test\/} set were generated by the organizers of the PAN challenge via the submitted program code.}. 
Rows 1-3 show the performance on the dev set and rows 6-8 show the corresponding results on the test set.
We used the {\em early-bird\/} feature of the challenge to get a first impression of how our model behaves on the test data.
The comparatively good results of our early-bird submission on the dev data (see row 1) suggest that our train and dev sets must be approximately stratified. Comparing these results with the significantly lower performance of the early-bird system on the test set (see row 6), however,
%
indicates that
there must be some type of intentional mismatch between the train set and the test set of the challenge. We suspect a shift in the relation between authors and fandom topics. For our early-bird submission we did not yet use the provided fandom labels. After the early-bird deadline, however, we incorporated the contextual prefixes. Comparing row~1 (without prefix) with row~4 (prefix included) we observe a
noticeable improvement. One possible explanation for this improvement could be that the model is now better able to recognize stylistic variations between authors who are writing in the same fandom-based domain. 
If we compare rows~4 \& 5 with rows~2 \& 3, we see the benefits of the proposed ensemble inference strategy. Combining a set of trained models leads to higher scores. Comparing rows 2 \& 3 and rows 7 \& 8 we find, unsurprisingly, that the training on the large dataset improves the  performance results as well.

Besides the losses in Eqs.~\eqref{eq:loss1} and~\eqref{eq:loss2}, we can also take into account the between-author and within-author variations to validate the training progress of our model. Both, between-author and within-author variations can be characterized by determining the entropy w.r.t. the estimated covariance matrices $\boldsymbol{\widehat{B}}^{-1}$ and $\boldsymbol{\widehat{W}}^{-1}$. It is  well-known that entropy can function as a measure of uncertainty. For multivariate Gaussian densities, the analytic solution of the  entropy is proportional to the determinant of the covariance matrix. From Eq.~\eqref{eg:Gaussian1} and~\eqref{eg:Gaussian2}, we have
\begin{align}
    H\big( \mathcal{N}(\boldsymbol{x} | \boldsymbol{\mu}, \boldsymbol{B}^{-1}) \big) 
            \propto  \log \det \boldsymbol{\widehat{B}}^{-1} 
    \quad \text{and} \quad
    H\big( \mathcal{N}(\boldsymbol{\epsilon} | \boldsymbol{0}, \boldsymbol{W}^{-1} \big) 
            \propto  \log \det \boldsymbol{\widehat{W}}^{-1}. 
\end{align}
Fig.~\ref{fig:en_cur} presents the entropy curves. As expected, during the training, the within-author variability decreased while the between-author variability increased.

Fig.~\ref{fig:example} shows the attention-heatmaps of two fanfiction excerpts. 
From a visual inspection of many of such heatmaps
we made the following observations: In contrast to the Amazon reviews used in~\cite{DBLP:conf/bigdataconf/BoenninghoffHKN19}, fanfiction texts do not contain a lot of "easy-to-visualize" linguistic features such as spelling errors for example. 
The model focuses on different aspects. Similar to~\cite{DBLP:conf/bigdataconf/BoenninghoffHKN19}, the model rarely marked function words (e.g. articles, pronouns, conjunctions). Surprisingly, punctuation marks like "\texttt{...}" seem to be less important than observed in~\cite{DBLP:conf/bigdataconf/BoenninghoffHKN19}.
In the first sentence of excerpt 1, the phrase "\texttt{stopped and looked}" is marked. In the second sentence, the word "\texttt{look}" of this phrase
is repeated in the overlapping part but not marked anymore. Contrarily, repeated single words like "\texttt{Absolutely}" in excerpt 2 remain marked. It seems that our model is able to analyze how an author is using a word in a particular context.

\begin{figure}[t!]
 \vspace*{-.15cm}
\centering
\footnotesize
\begin{testexample}
        \scalebox{0.85}{
        \input{pics/doc}
        }
\end{testexample}
\vspace*{-0.2cm}
\begin{testexample}
        \scalebox{0.85}{
            \input{pics/doc2}
            }
\end{testexample}
\caption{Attention-heatmaps. Blue hues encode the sentence-based attention weights and red hues denote the relative word importance. All tokens are delimited by whitespaces. 
\vspace*{-.5cm}}
\label{fig:example}
\end{figure}

Lastly, to keep the CPU memory requirements as low as possible on Tira~\cite{tira:2019}, we fed every single test document separately and
sequentially into the ensemble of trained models, resulting in a runtime of approximately 6 hours. This can, of course, be done batch-wise and in parallel for all models in the ensemble to reduce training time.

%% file: pics/doc.tex
{\setlength{\fboxsep}{0pt}\colorbox{white!0}{\parbox{13.7cm}{
\colorbox{blue!57.79708}{\strut ~~~}\colorbox{red!0.99822366}{\strut <Harry Potter>} \colorbox{red!2.3433888}{\strut grabbed} \colorbox{red!3.0517106}{\strut Scarlet} \colorbox{red!2.325471}{\strut ,} \colorbox{red!4.361297}{\strut and} \colorbox{red!10.242787}{\strut rushed} \colorbox{red!2.985813}{\strut out} \colorbox{red!1.571575}{\strut of} \colorbox{red!1.4901577}{\strut the} \colorbox{red!14.269805}{\strut common} \colorbox{red!4.276272}{\strut room} \colorbox{red!1.6036623}{\strut .} \colorbox{red!1.2543491}{\strut '} \colorbox{red!5.7331266}{\strut Draco} \colorbox{red!3.1702883}{\strut ,} \colorbox{red!3.9537165}{\strut let} \colorbox{red!2.3398652}{\strut go} \colorbox{red!1.6893005}{\strut you} \colorbox{red!2.831534}{\strut 're} \colorbox{red!7.8963943}{\strut hurting} \colorbox{red!2.2092378}{\strut me} \colorbox{red!2.2355628}{\strut .} \colorbox{red!2.2770228}{\strut '} \colorbox{red!4.2369404}{\strut He} \colorbox{red!23.472609}{\strut stopped} \colorbox{red!10.555504}{\strut and} \colorbox{red!23.242208}{\strut looked} \colorbox{red!3.4963675}{\strut at} \colorbox{red!3.382076}{\strut her} \colorbox{red!44.552578}{\strut bruising} 
 \newline\colorbox{blue!100.0}{\strut ~~~}\colorbox{red!0.93651724}{\strut <Harry Potter>} \colorbox{red!3.349636}{\strut looked} \colorbox{red!0.6260337}{\strut at} \colorbox{red!0.6448256}{\strut her} \colorbox{red!17.990814}{\strut bruising} \colorbox{red!1.4510235}{\strut wrist} \colorbox{red!0.20937866}{\strut .} \colorbox{red!0.12468455}{\strut '} \colorbox{red!4.270505}{\strut Sorry} \colorbox{red!0.5304165}{\strut ,} \colorbox{red!0.09694571}{\strut '} \colorbox{red!0.49061748}{\strut he} \colorbox{red!4.271533}{\strut said} \colorbox{red!6.594303}{\strut letting} \colorbox{red!1.2653924}{\strut go} \colorbox{red!0.62832993}{\strut .} \colorbox{red!1.168832}{\strut She} \colorbox{red!29.397661}{\strut rubbed} \colorbox{red!0.8221917}{\strut it} \colorbox{red!1.9165272}{\strut ,} \colorbox{red!52.168346}{\strut hissing} \colorbox{red!1.1473496}{\strut a} \colorbox{red!16.939507}{\strut bit} \colorbox{red!1.6003433}{\strut at} \colorbox{red!1.6009642}{\strut the} \colorbox{red!47.03682}{\strut soreness} \colorbox{red!0.0}{\strut .} \colorbox{red!0.025968315}{\strut '} \colorbox{red!0.423858}{\strut It} \colorbox{red!0.31956646}{\strut 's} 
 \newline\colorbox{blue!36.60667}{\strut ~~~}\colorbox{red!0.96425015}{\strut <Harry Potter>} \colorbox{red!0.8189451}{\strut .} \colorbox{red!0.8463207}{\strut '} \colorbox{red!1.2926078}{\strut It} \colorbox{red!1.2180892}{\strut 's} \colorbox{red!25.245127}{\strut alright} \colorbox{red!1.6516387}{\strut .} \colorbox{red!3.2328691}{\strut Why} \colorbox{red!2.2146184}{\strut were} \colorbox{red!1.7127262}{\strut you} \colorbox{red!16.047626}{\strut arguing} \colorbox{red!2.5415092}{\strut with} \colorbox{red!3.637214}{\strut them} \colorbox{red!1.5099461}{\strut in} \colorbox{red!1.3422557}{\strut the} \colorbox{red!3.2396655}{\strut first} \colorbox{red!5.595361}{\strut place} \colorbox{red!1.4593747}{\strut ?} \colorbox{red!1.7304381}{\strut '} \colorbox{red!3.0545094}{\strut He} \colorbox{red!48.13146}{\strut hesitated} \colorbox{red!4.682803}{\strut for} \colorbox{red!2.2626567}{\strut a} \colorbox{red!25.26162}{\strut moment} \colorbox{red!6.578555}{\strut and} \colorbox{red!20.645971}{\strut answered} \colorbox{red!2.4065113}{\strut ,} \colorbox{red!1.5078572}{\strut '} \colorbox{red!2.724608}{\strut She} \colorbox{red!4.4917455}{\strut just} 
}}}

%% file: pics/doc2.tex
{\setlength{\fboxsep}{0pt}\colorbox{white!0}{\parbox{13.7cm}{
\colorbox{blue!53.165447}{\strut ~~~}\colorbox{red!1.608013}{\strut <Batman>} \colorbox{red!1.9940386}{\strut with} \colorbox{red!1.3322556}{\strut an} \colorbox{red!3.4338996}{\strut update} \colorbox{red!1.0918832}{\strut as} \colorbox{red!2.2533445}{\strut soon} \colorbox{red!1.1733356}{\strut as} \colorbox{red!0.96877265}{\strut I} \colorbox{red!1.5718858}{\strut can} \colorbox{red!1.0783166}{\strut !} \colorbox{red!0.9101757}{\strut '} \colorbox{red!8.408094}{\strut Can} \colorbox{red!1.6685003}{\strut you} \colorbox{red!6.4647017}{\strut believe} \colorbox{red!2.3967156}{\strut that} \colorbox{red!1.3365703}{\strut ?} \colorbox{red!1.0981815}{\strut '} \colorbox{red!11.635331}{\strut Carly} \colorbox{red!2.8700814}{\strut was} \colorbox{red!55.50057}{\strut fuming} \colorbox{red!1.6063406}{\strut .} \colorbox{red!1.229265}{\strut '} \colorbox{red!31.331465}{\strut Fourteen} \colorbox{red!8.9646225}{\strut boys} \colorbox{red!3.0625646}{\strut in} \colorbox{red!3.7538047}{\strut one} \colorbox{red!3.3255248}{\strut house} \colorbox{red!2.575637}{\strut ?} \colorbox{red!20.350878}{\strut Absolutely} \colorbox{red!2.2859876}{\strut archaic} 
 \newline\colorbox{blue!21.24888}{\strut ~~~}\colorbox{red!3.1562328}{\strut <Batman>} \colorbox{red!3.197354}{\strut house} \colorbox{red!2.8688912}{\strut ?} \colorbox{red!36.99535}{\strut Absolutely} \colorbox{red!2.9863489}{\strut archaic} \colorbox{red!2.1203296}{\strut !} \colorbox{red!5.9313216}{\strut There} \colorbox{red!2.4786706}{\strut 's} \colorbox{red!11.775888}{\strut hardly} \colorbox{red!7.772222}{\strut enough} \colorbox{red!3.14553}{\strut room} \colorbox{red!3.202898}{\strut for} \colorbox{red!6.2722764}{\strut five} \colorbox{red!2.7414687}{\strut ,} \colorbox{red!5.241764}{\strut maybe} \colorbox{red!13.117999}{\strut ten} \colorbox{red!2.728771}{\strut ...} \colorbox{red!6.642392}{\strut And} \colorbox{red!3.630533}{\strut his} \colorbox{red!10.237481}{\strut eye} \colorbox{red!2.8181548}{\strut !} \colorbox{red!4.3459735}{\strut How} \colorbox{red!6.937442}{\strut could} \colorbox{red!3.6946378}{\strut they} \colorbox{red!5.1700144}{\strut let} \colorbox{red!4.0821576}{\strut that} \colorbox{red!11.800795}{\strut happen} \colorbox{red!3.2394097}{\strut ?} \colorbox{red!4.931882}{\strut Oh} \colorbox{red!4.016535}{\strut ,} 
 \newline\colorbox{blue!38.87512}{\strut ~~~}\colorbox{red!2.45359}{\strut <Batman>} \colorbox{red!5.4969687}{\strut happen} \colorbox{red!2.785423}{\strut ?} \colorbox{red!4.7454195}{\strut Oh} \colorbox{red!4.1914277}{\strut ,} \colorbox{red!2.5762746}{\strut I} \colorbox{red!9.011795}{\strut know} \colorbox{red!2.149097}{\strut !} \colorbox{red!5.721598}{\strut there} \colorbox{red!2.8970563}{\strut 's} \colorbox{red!3.1734734}{\strut a} \colorbox{red!16.72172}{\strut half} \colorbox{red!43.660248}{\strut dozen} \colorbox{red!7.34128}{\strut kids} \colorbox{red!6.38692}{\strut too} \colorbox{red!4.9515314}{\strut many} \colorbox{red!15.1933775}{\strut occupying} \colorbox{red!2.0724666}{\strut a} \colorbox{red!3.0055702}{\strut space} \colorbox{red!3.84329}{\strut for} \colorbox{red!3.0015655}{\strut ...} \colorbox{red!9.691216}{\strut Are} \colorbox{red!2.810974}{\strut you} \colorbox{red!4.1446686}{\strut even} \colorbox{red!6.9983096}{\strut listening} \colorbox{red!2.2202759}{\strut to} \colorbox{red!2.305931}{\strut me} \colorbox{red!1.9616365}{\strut ?} \colorbox{red!2.4055886}{\strut '} \colorbox{red!3.3620827}{\strut She} 
}}}

%% file: sections/sec_5_conclusion.tex
We presented a new type of authorship verification (AV) system that combines neural feature extraction with statistical modeling. By recombining document-pairs after each training epoch, we significantly increased the heterogeneity of the train data. %
%
The proposed method achieved excellent overall performance scores, outperforming all other systems that participated in the PAN 2020 Authorship Verification Task, in both the small dataset challenge as well as the large dataset challenge.
%
%
%
In AV there are many variabilities (such as topic, genre, text length, etc.) that negatively affect the system performance.
Great opportunities for further gains can, thus, be expected by incorporating \textit{compensation techniques} that deal with these aspects in future challenges.

%% file: main.bbl
\begin{thebibliography}{10}
\providecommand{\url}[1]{\texttt{#1}}
\providecommand{\urlprefix}{URL }

\bibitem{articlebagnall}
Bagnall, D.: {Author Identification using multi-headed Recurrent Neural
  Networks}. In: {CLEF Evaluation Labs and Workshop -- Working Notes Papers}
  (2015)

\bibitem{Bahdanau14}
Bahdanau, D., Cho, K., Bengio, Y.: {Neural Machine Translation by Jointly
  Learning to Align and Translate}. In: Proc. {ICLR} (2015)

\bibitem{bischoff:2020}
Bischoff, S., Deckers, N., Schliebs, M., Thies, B., Hagen, M., Stamatatos, E.,
  Stein, B., Potthast, M.: {The Importance of Suppressing Domain Style in
  Authorship Analysis}. CoRR  abs/2005.14714 (2020)

\bibitem{DBLP:conf/bigdataconf/BoenninghoffHKN19}
Boenninghoff, B., Hessler, S., Kolossa, D., Nickel, R.M.: {Explainable
  Authorship Verification in Social Media via Attention-based Similarity
  Learning}. In: Proc. {IEEE} BigData (2019)

\bibitem{HRSN}
Boenninghoff, B., Nickel, R.M., Zeiler, S., Kolossa, D.: {Similarity Learning
  for Authorship Verification in Social Media}. In: Proc. ICASSP (2019)

\bibitem{niko220}
{N}iko {B}r{\"u}mmer, {E}dward~de {V}illiers: {The speaker partitioning
  problem}. In: Proc. Odyssey. ISCA (2010)

\bibitem{rummer1111}
Br{\"u}mmer, N.: {A farewell to SVM: Bayes factor speaker detection in
  supervector space}. Tech. rep. (2006)

\bibitem{DBLP:journals/corr/abs-2003-11982}
Chung, J.S., Huh, J., Mun, S., Lee, M., Heo, H.S., Choe, S., Ham, C., Jung, S.,
  Lee, B., Han, I.: {In defence of metric learning for speaker recognition}.
  CoRR  abs/2003.11982 (2020)

\bibitem{6466371}
{Cumani}, S., {Br{\"u}mmer}, N., {Burget}, L., {Laface}, P., {Plchot}, O.,
  {Vasilakakis}, V.: {Pairwise Discriminative Speaker Verification in the
  I-Vector Space}. {IEEE Trans. Audio, Speech, Lang. Process.}  (2013)

\bibitem{GVK021834997}
DeGroot, M.: {Optimal statistical decisions}. McGraw-Hill (1970)

\bibitem{BKA}
Ehrhardt, S.: Authorship attribution analysis. In: Visconti, J. (ed.) Handbook
  of Communication in the Legal Sphere. pp. 169--200. de Gruyter, Berlin/Boston
  (2018)

\bibitem{HalvaniUnaryBinary}
Halvani, O., Winter, C., Graner, L.: {Assessing the Applicability of Authorship
  Verification Methods}. In: Proc. ARES (2019)

\bibitem{LSTM}
Hochreiter, S., Schmidhuber, J.: {Long Short-Term Memory}. Neural Comp.  (1997)

\bibitem{Hu14}
Hu, J., Lu, J., Tan, Y.P.: {Discriminative Deep Metric Learning for Face
  Verification in the Wild}. In: Proc. CVPR (2014)

\bibitem{10.1007/11744085_41}
Ioffe, S.: {Probabilistic Linear Discriminant Analysis}. In: Leonardis, A.,
  Bischof, H., Pinz, A. (eds.) Proc. {ECCV} (2006)

\bibitem{kestemont:2020}
Kestemont, M., Manjavacas, E., Markov, I., Bevendorff, J., Wiegmann, M.,
  Stamatatos, E., Potthast, M., Stein, B.: {Overview of the Cross-Domain
  Authorship Verification Task at PAN 2020}. In: Cappellato, L., Eickhoff, C.,
  Ferro, N., N{\'e}v{\'e}ol, A. (eds.) {CLEF 2020 Labs and Workshops, Notebook
  Papers}. CEUR-WS.org (2020)

\bibitem{DBLP:conf/simbig/Litvak18}
Litvak, M.: {Deep Dive into Authorship Verification of Email Messages with
  Convolutional Neural Network}. In: Proc. SIMBig (2018)

\bibitem{phdthesis}
Potha, N.: Authorship Verification. Ph.D. thesis, University of the Aegean
  (2019)

\bibitem{tira:2019}
Potthast, M., Gollub, T., Wiegmann, M., Stein, B.: {TIRA Integrated Research
  Architecture}. In: Ferro, N., Peters, C. (eds.) {IR Evaluation in a Changing
  World}. Springer (2019)

\bibitem{Stamatatos09}
Stamatatos, E.: {A Survey of Modern Authorship Attribution Methods}. J. Assoc.
  Inf. Sci. Technol.  (2009)

\bibitem{stamatatos-2017-authorship}
Stamatatos, E.: {Authorship Attribution Using Text Distortion}. In: Proc. EACL
  (2017)

\end{thebibliography}
